\documentclass[conference]{IEEEtran}
\usepackage{acro}
\usepackage[utf8]{inputenc}
\usepackage[ruled,vlined,linesnumbered]{algorithm2e}
\usepackage{amsfonts, amsmath, amssymb}
\usepackage{array}
\usepackage{bm}
\usepackage{cite}
\usepackage{color}
\usepackage{float}
\usepackage{subfig}
\usepackage{hyperref}
\usepackage{url}
\usepackage{multirow}
\usepackage{booktabs}
\usepackage{graphicx}
 \usepackage[switch]{lineno}
\usepackage{makecell}

\title{Wireless teleoperation of HSURF artificial fish in complex paths}

\author{
\IEEEauthorblockN{ Saverio Iacoponi} %2\textsuperscript{nd}
\IEEEauthorblockA{\textit{Dept. of Mechanical Eng.} \\
\textit{Khalifa University}\\
Abu Dhabi, UAE\\
}

\and

\IEEEauthorblockN{ Nikita Mankovskii} %3\textsuperscript{rd}
\IEEEauthorblockA{\textit{ARRC} \\
\textit{Technology Innovation Institute}\\
Abu Dhabi, UAE \\
}
\and

\IEEEauthorblockN{ Mohammed El Hanbaly} %2\textsuperscript{nd}
\IEEEauthorblockA{\textit{Dept. of Mechanical Eng.} \\
\textit{Khalifa University}\\
Abu Dhabi, UAE\\
}

\and 

\IEEEauthorblockN{ Andrea Infanti} %2\textsuperscript{nd}
\IEEEauthorblockA{\textit{Dept. of Mechanical Eng.} \\
\textit{Khalifa University}\\
Abu Dhabi, UAE\\
}

\and 
\IEEEauthorblockN{ Shamma Alhajeri} %6\textsuperscript{th}
\IEEEauthorblockA{\textit{ARRC} \\
\textit{Technology Innovation Institute}\\
Abu Dhabi, UAE \\
}

\and
\IEEEauthorblockN{ Federico Renda} %4\textsuperscript{th}
\IEEEauthorblockA{\textit{Dept. of Mechanical Eng.} \\
\textit{Khalifa University}\\
Abu Dhabi, UAE\\
}
\and
\IEEEauthorblockN{ Cesare Stefanini} %5\textsuperscript{th}
\IEEEauthorblockA{\textit{BioRobotics Institute} \\
\textit{Scuola Superiore Sant'Anna}\\
Pisa, Italy\\
}
\and 
\IEEEauthorblockN{ Giulia De Masi} %1\textsuperscript{th}
\IEEEauthorblockA{
\textit{ Technology Innovation Institute}\\
\textit{Khalifa University}\\
Abu Dhabi, UAE \\
}

}

\begin{document}
    \maketitle
\begin{abstract}
In this paper we show the application of the new robotic multi-platform system HSURF to a specific use case of teleoperation,  aimed at monitoring and inspection. The HSURF system, consists of 3 different kinds of platforms: floater, sinker and robotic fishes. The collaborative control of the 3 platforms allows a remotely based operator to control the  fish in order to visit and inspect several targets underwater  following a complex trajectory. A shared autonomy solution shows to be the  most suitable, in order to minimize the effect of limited bandwidth and relevant delay intrinsic to acoustic communications. The control architecture is described and preliminary results of the acoustically teleoperated visits of multiple targets in a testing pool are provided. 
\end{abstract}

% The preservation of the health status of the Ocean is of primary importance. The program ``Decade of Ocean Science for Sustainable Development'' (2021-2030) has been declared by United Nation \cite{oceandecade} in order to activate all the stakeholders playing a role in the marine environment to invert the trend of decline of the health status of the oceans and raise sensibility on the preservation and restoration of the underwater environment. Undoubtedly, the oceans are more and more exploited for commercial purposes, from shipping routes, to wind farms, oil and gas facilities, civil structures, mineral extractions, fish farms and algae cultivation. 
% As a consequence, many threads are affecting the natural underwater environment: from plastic circulation to materials' dispersion, to oil and gas structures leakages and consequent  oil spills. In this sense, continuous monitoring and inspection play a crucial role to avoid any environmental crisis. Also surveillance of fish farms, ship hulls, civil and recreational structures are important to guarantee the proper functioning of all these facilities. In order to perform regular inspection and surveillance, untethered underwater vehicles are ideal, mostly when these activities are performed in complex structured environments, like harbors, piers, archaeological sites, oil\&gas facilities\cite{sudevan2022multisensor}. 

% COMPACTED INTRO

\section{Introduction}

The preservation of the health status of the Ocean is of primary importance. To this aim, the program "Decade of Ocean Science for Sustainable Development" (2021-2030) has been declared by United Nation \cite{oceandecade}. Undoubtedly, the oceans are more and more exploited for commercial purposes, and, as a consequence, many threads are affecting the natural underwater environment: from plastic circulation to materials' dispersion, to oil\&gas structures' leakages and consequent  oil spills. Continuous surveillance and inspections of hulls and seabed infrastructures become fundamental. In order to perform regular inspection and surveillance, untethered underwater vehicles are ideal, mostly when these activities are performed in complex structured environments, like harbors, piers, archaeological sites, oil\&gas facilities\cite{sudevan2022multisensor}. 

\begin{figure}[!ht]
\includegraphics[width=7.5cm,height=4cm]{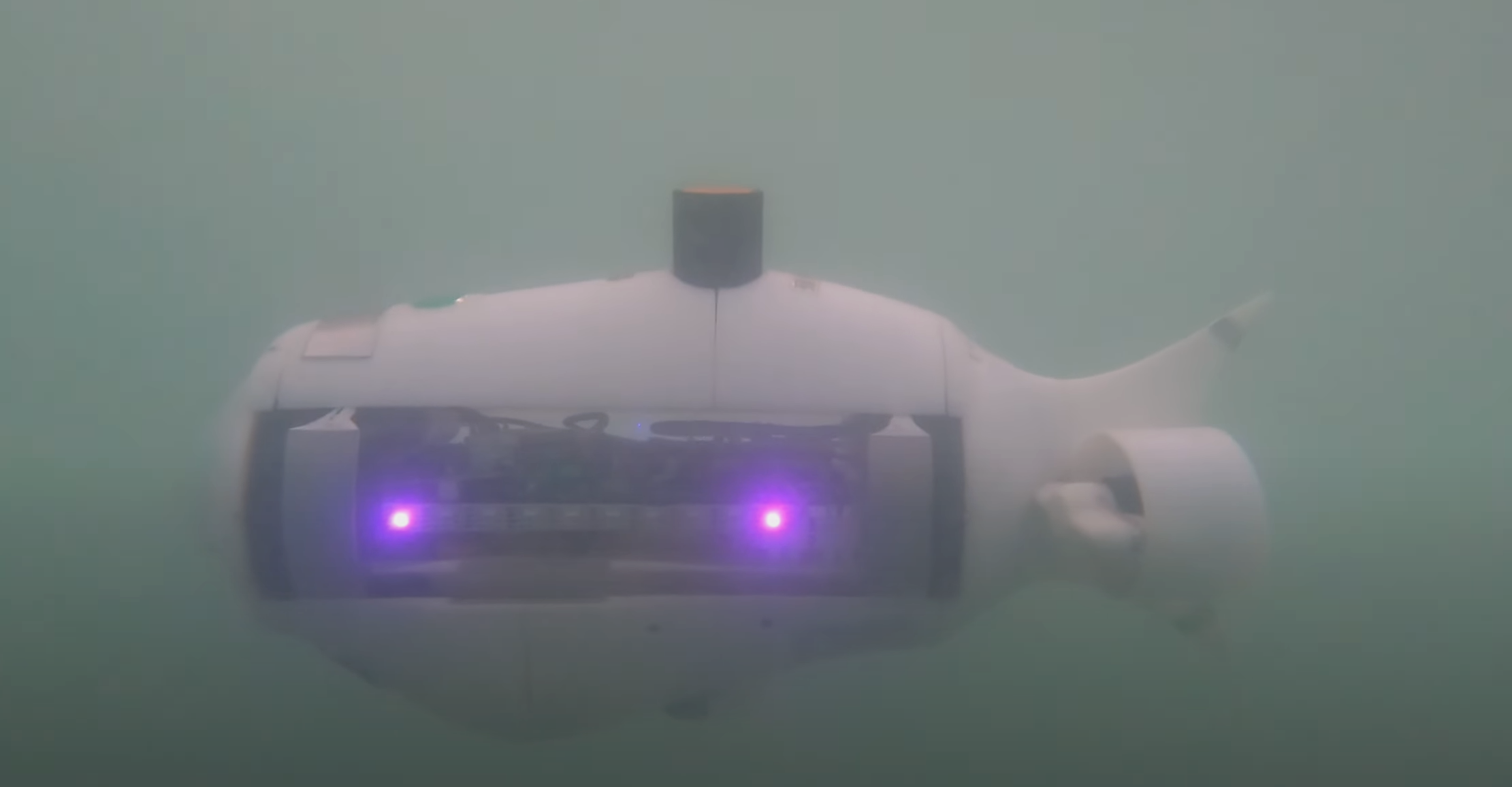}
\centering
\caption{HSURF fish picture during preliminary open-water tests.}
\label{fig:fish}
\end{figure}

The HSURF platform system has been designed and developed also for monitoring and surveillance missions, as already described in \cite{iacoponi2022h}: an integrated system consisting of a floater platform, a sinker and 20 artificial fishes (AUVs). A picture of the artificial fish, captured during qualitative testing of teleoperation on open waters, is presented in Fig.\ref{fig:fish}. As presented in the \cite{iacoponi2022iop}, the control and coordination of the swarm of fish relies on combining single AUV autonomy, with coordination between AUVs and operator feedback. 

With focus on the last aspect, multi-modal communication and a smart use of limited underwater bandwidth is required to handle communication between the operator and the AUV without relying to a tethered system \cite{app10041256}.
Wireless control of underwater vehicles have been discussed at length in \cite{Campagnaro2015SimulationOA,7094391,jmse8100736},  and it is clear that the teleoperation topic is of great interest in the community \cite{muscolo_marcheschi_fontana_bergamasco_2021} for a range of applications including  manipulation in an augmented reality setting \cite{Cardenas2016}, with increasing focus more and more on optimal human-robot interfaces \cite{padrao2023towards}.  
Acoustic-based remote control is particularly challenging due to the main drawbacks of acoustic communication: limited bandwidth and considerable delay \cite{Akyildiz2004}.  

Here we use the HSURF multi-platforms' system to a remotely controlled multi-target inspection and monitoring. This use case is particularly relevant for subsea structures' inspections, where many components (valves, pipes, risers) need to be controlled during the same mission.

In this paper, we are presenting experimental testing of wireless teleportation based on the acoustic channel of a single HSURF robotic  fish by a human operator. We showcase preliminary results of remote teleoperation by presenting quantitative and qualitative analyses of the performance obtainable with the proposed method. 

\section{Teleportation Architecture and Goals}

The general idea is that an operator is able to  perform a survey mission at open sea, being able to control the robot using teleoperation, in order to  inspect the underwater biome and structures, and collect information while recording videos, in a data-muling scenario. 

From visual inspection,  the operator can assess the status of underwater structures, like  the integrity of civil structures, or  the level of corrosion of the  oil\&gas facilities, as well as the presence of leaks, or  the overall health status of natural ecosystems (like coral reefs).

The first requirement for the envisioned operation is for the operator to be able to control the robot underwater. However, a limited and delayed communication does not allow for direct actuation control, while the AUV relies only on limited sensing: pressure (used for depth estimation) and a 9DoF IMU including a  magnetometer (used for heading estimation and rotation rate). 

Direct control of the actuators is not possible due to the extremely limited and delayed nature of the acoustic channel. Therefore, a certain level of autonomy must be granted to the AUV itself. Instead of directly sending motors commands, we opted for commanding a desired state, in terms of heading and depth. The AUV then uses its onboard sensors to autonomously control the actuation and reach the desired state. 

During the teleoperation, desired state (depth, heading and forward-backward thrust) is given to the robot based on the operator inputs. To allow for the fastest rate of transmission, the three information are compressed in a single byte. Heading and forward-backward are quantized in 16 cardinal directions and 5 states respectively (back, slow-back, stop, slow-forward, forward). The depth is controlled in increments, with 3 states (lower 2.5cm, hold depth, higher 2.5cm). The total of 240 combinations is then encoded in 1 byte. The byte is communicated from the operator to the "floater"  via WiFi, from which is broadcasted underwater by an acoustic modem (Succorfish V2). 

 Actual depth, rotation rate and heading, obtained by the onboard sensors, are used  by PID controllers, along with the last desired state received. The controllers return outputs as vertical thrust $F_z$ and torque around the vertical axis $M_z$. The depth PID controller directly takes the depth as input, while for the heading we adopted a cascade controller: a proportional controller outputs a desired rotation rate, which is input in a PID with the rotation rate of the onboard IMU gyroscope as feedback. $F_x$ is instead controlled in open-loop, since the AUV does not have any mean of measuring its own velocity or position. Using the vehicle Thrust Allocation Matrix, those are converted in motor thrusts, and then in motor commands \cite{fossen2011handbook}. 

\section{Method}
\begin{figure}[!ht]
\includegraphics[width=7.5cm,height=4cm]{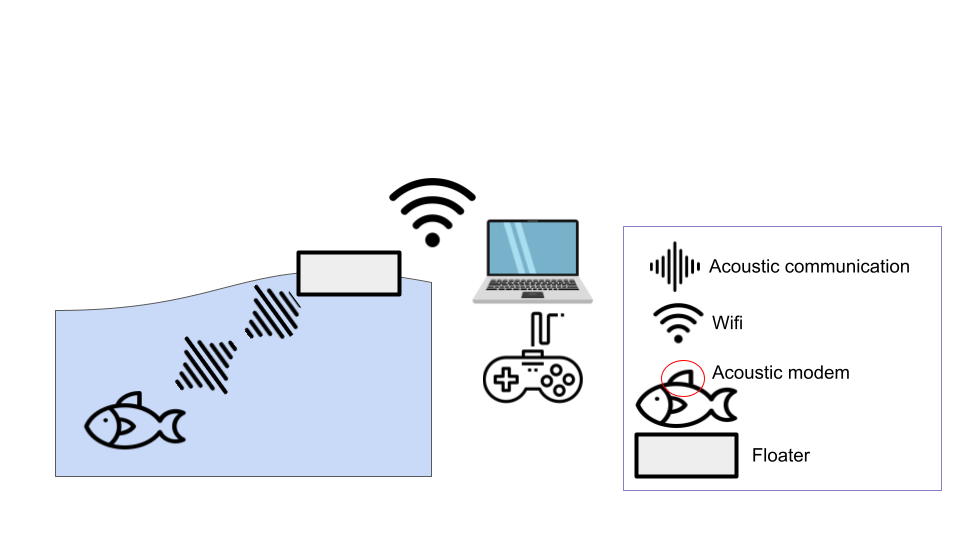}
\centering
\caption{Sketch of the experiment configuration.}
\label{fig:sketch}
\end{figure}

 The AUV architecture and control system are described in depth in \cite{iacoponi2022h}. Of particular interest for this work is the HSURF fish-robot. The fish-robot in our testing is a 4kg, 450mm long underwater autonomous underwater vehicle. It features a 9-DoF IMU, an V2 Succorfish acoustic modem, and 3 propellers (1 for vertical motions and 2 for planar motion). The HSURF fish is obviously under-actuated and allows for direct control of surge and heading but not sway. The AUV is powered by its own batteries and controlled by a RaspberryPi4 on which it is executed the control software, which is based on ROS-Noetic. 
The system is sketched in Fig.\ref{fig:sketch}. The teleoperation is based on acoustic communication. 

The remote control happens through a joystick controller connected via Bluetooth to the PC. The PC sends the signal through WiFi to the platform floating on the surface. For the current test, we used a simple platform including  a microprocessor and an acoustic modem. The signal is encoded and re-transmitted at regular intervals, using acoustic communication which is propagating underwater. Only the most updated joystick communication is sent for every communication slot. For the current configuration, a frequency of 1 message per 1.6 seconds is achieved, which is limited by the acoustic modem proprieties and pool echo. The underwater artificial fish receives and interprets the message, acting accordingly to the instruction received until a new message is obtained. The internal control system allows to maintain control of the AUV, with limited command adjustment, then can be transmitted acoustically.

Both surface and underwater AUVs are based on ROS software architecture, although they are isolated from each other when underwater.

%By relying in ROS-Gazebo, we are able to simulate the full dynamics of the underwater AUV. A simulator was developed to closely emulate the dynamics of an AUVs, actuation and hydrodynamic effects. The simulation allows for real time simulations, and we have included a node that introduces the acoustic communication delay and bottle-necking of transmittable data. The dynamic control and signal encoding and decoding are run in ROS-Noetic, which acquires the input of a joystick for controls. A user can then operate the simulated robot via the GUI interface of Gazebo. Commands from joystick are quantized and encoded, pass through the modem emulator for delay and then given to the simulated AUV control system. We added some visual markers (1 meters diameters spheres), acting as waypoints for the user. Figure \ref{fig:path} shows the simulated AUV path and the target markers. This proves that the architecture allows to compensate for the delay and bottlenecking effect of the acoustic communication channel. 

% \begin{figure}[!ht]
% \includegraphics[width=7.0cm]{figures/path.jpg}
% \centering
% \caption{Simulated test. The operator moved the AUV through all six targets, represented in red circles. The blue path represent the resulting trajectory. Delays, dynamic and data packaging emulated the real case scenario. }
% \label{fig:path}
% \end{figure}

\section{Tests}

\begin{figure}[!ht]
\includegraphics[width=7.0cm]{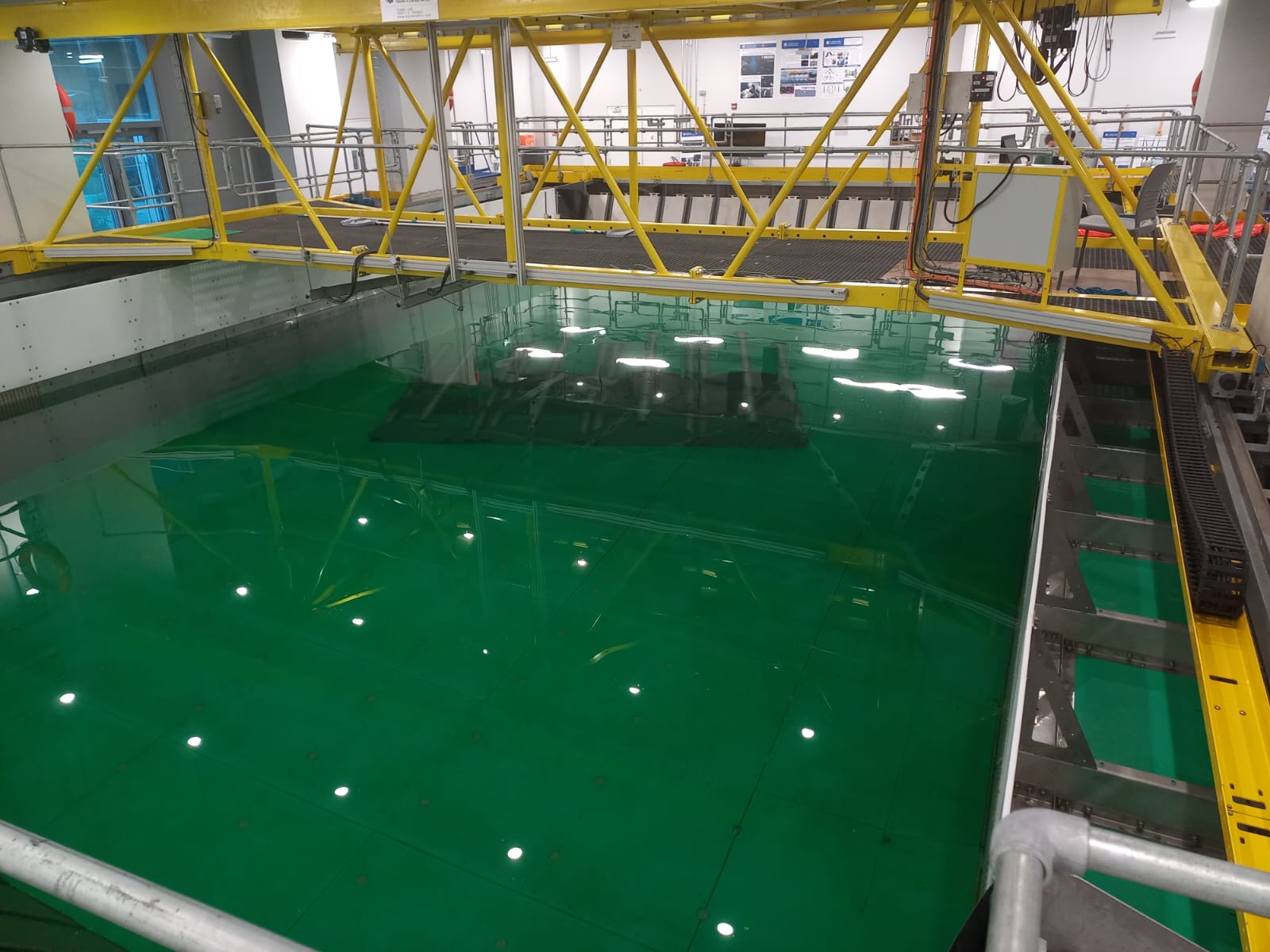}
\centering
\caption{Testing Pool, Marine robotics pool in Khalifa University.}
\label{fig:pool}
\end{figure}

The tests were performed in the marine robotics pool in Khalifa University. The pool, in Fig.\ref{fig:pool}, is 12.5m by 8m by 2.1m. The pool's walls host the waves and current generators and were kept off during testing. A previous verification showcased that while acoustic communication is possible in the water, the signal quality is heavily degraded by the reverberation effect, and de-facto pushes the acoustic to operate close to its limits of operablility. 

At the beginning of the day, magnetic calibration of the magnetometer is performed in the pool. As the pool bottom and wall are for the most part made of stainless steel, and the pool is in an enclosed area, the magnetic field is affected. Nonetheless, the magnetic field is constant enough to be usable as absolute reference.

\begin{figure}[!ht]
\includegraphics[width=7.0cm]{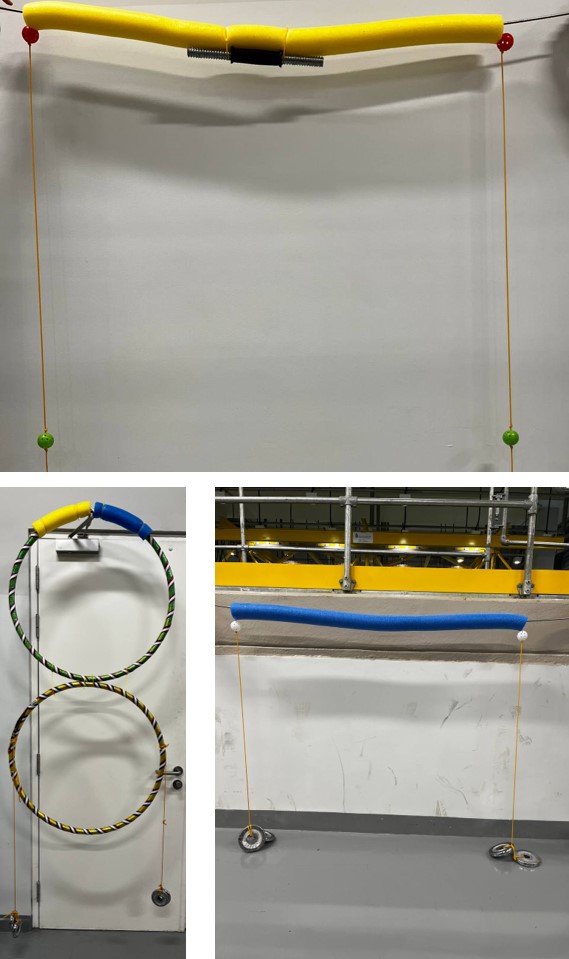}
\centering
\caption{Picture of the target-gates. Top panel:  the gate 1, with the floating foam on the surface and 2 markers delimiting the bottom limit at a depth of 1m and a width of 1m. Bottom left panel:  gates 2 and 4 which have a diameter of 0.75m, with gate 2 reaching the water surface. Bottom right panel: gate 3, with a full height of 0.9m and 1.1m width}
\label{fig:gates1}
\end{figure}

\begin{figure}[!ht]
\includegraphics[width=8.0cm]{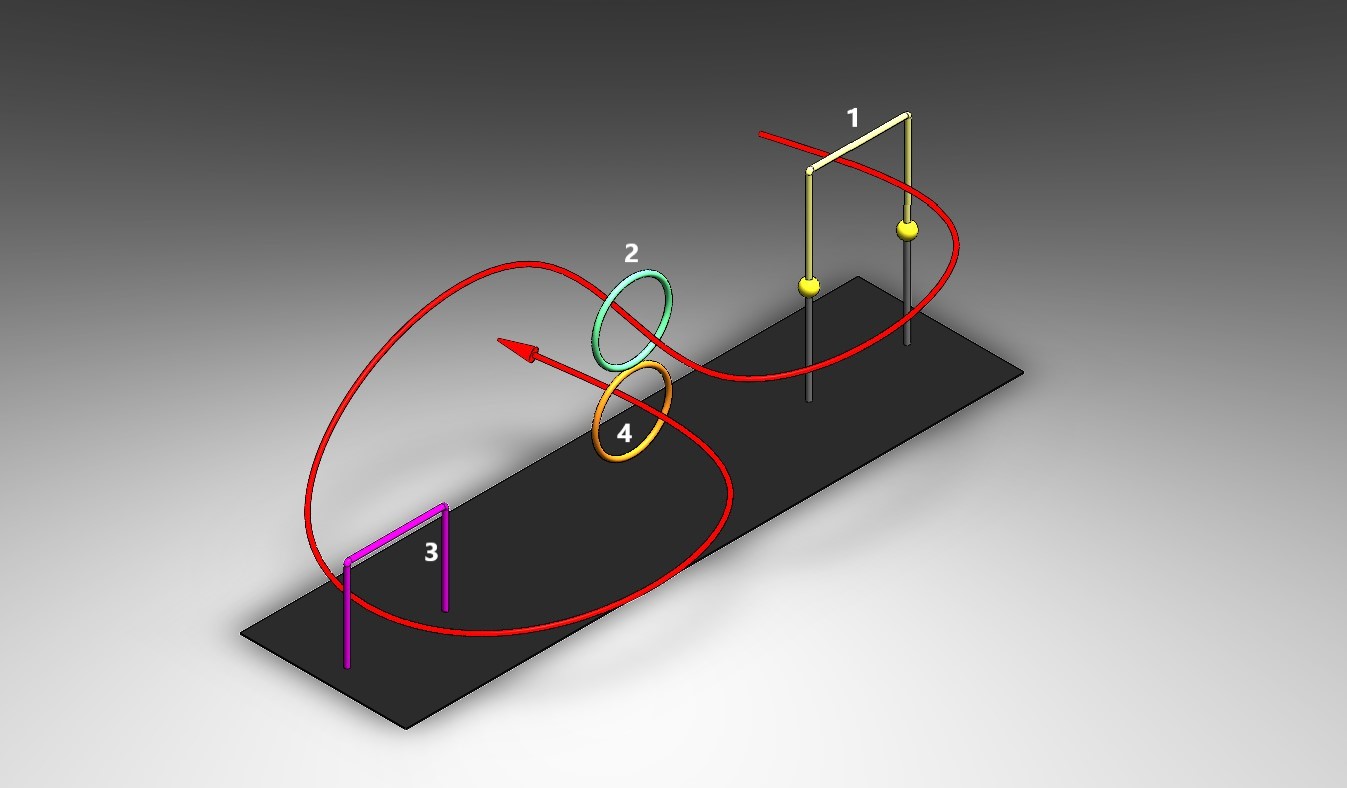}
\centering
\caption{Schematic of a test. Gates 1, 2\&4 and 3 are respectively placed on a line at approximately 2 meters distance each. The robot enters the gate 1, maintaining a depth between 1m and the surface, it performs a U-Turn and threads through gate 2. It then performs a second U-turn and descends to pass through gate 3. Finally, it raises and U-turns to pass through gate 4.}
\label{fig:infograph}
\end{figure}

A set of targets were placed in the pool, positioned as shown in Fig.\ref{fig:gates1}. The target consists in  vertical gates through which the AUV has to pass through, as shown in Fig.\ref{fig:infograph}. The first target is positioned 1 meter below the surface, with visible markers to delimit the maximum allowed depth, second and fourth targets are circular gates of 0.75m diameter placed at different depths, one above the other, while the third target extends 0.90m from the bottom. The AUV must pass each gate in the right order from the desired side: the AUV is teleoperated in front of the first gate, then the starting time is taken, the fish passes the first gate, performs a U turn, passes the second gate, performs a second U turn and reducing the depth of approximately 1 meter pass the third gate, finally cover the last U turn and does a small upward adjustment of depth to pass the fourth gate, when  the test is ended.

The test was repeated 5 times. For each test the fish passed through the 4 gates. We recorded the total time of each test (counted from the first attempt at gate 1 to the successful passing of gate 4), the number of attempts to pass each gate, as well as the list of sent and received messages over time. Note that we ensured the PC and fish clocks to be synchronized. If any attempt to pass a gate resulted in failure, the vehicle was teleoperated back and the attempt repeated. The gate has to be traversed only from the intended side and the time count was not interrupted. 

\section{Results}

Using the recorded ROS bags on both PC and AUV, we could extract further information regarding the communication. Firstly a statistic on the effective time delay between each sent message confirmed that the time distance between transmission was 1.6 seconds on average, with minimal variation between each test. 

By analysing the command sent and received, we extracted a statistics of message loss rate and effective communication delay.

The message loss rate is calculated on commands variation (we consider a command variation when the new encoded command differs from the previous command sent). The total number of command variations per test is reported as well as the percentage of messages lost during command variations. The counting of command losses is made over command variations, as they are the ones affecting the dynamics of the system, reducing the responsiveness of the AUV and introducing an occasional control delay. 

Time delay is measured from the moment the message is sent to the modem for transmission (from the PC), to the execution of the command itself on the HUSRF-fish. This includes WiFi transmission delay, acoustic delay and all processing and structural delays. For each test, the average delay and variance are reported.

A full table of the results is presented in Tab.\ref{table_attempts} and Tab.\ref{table_statistics}.

\begin{table}[htbp]
\caption{Tests' Results: Number of attempts per gate}
\begin{center}
\begin{tabular}{|c||c|c|c|c|c|}
\hline
\textbf{Test N.}  & \thead{Total \\ test\\ time [s]} & \thead{ Gate 1 \\Attempts} & \thead{ Gate 2 \\Attempts} & \thead{ Gate 3 \\Attempts} & \thead{ Gate 4 \\Attempts}  \\
\hline
Test 1& 219 &2 &1 &1 &1\\
Test 2& 129 &1 &1 &1 &1\\
Test 3& 329 &1 &2 &1 &3\\
Test 4& 181 &1 &1 &1 &1\\
Test 5& 133 &1 &1 &1 &1\\
\hline
\end{tabular}
\label{table_attempts}
\end{center}
\end{table}

\begin{table}[htbp]
\caption{Statistics on acoustic communication messages $^*$}
\begin{center}
\begin{tabular}{|c||c|c|c|c|}
\hline
\textbf{Test N.}  &   \thead{Message \\ loss percentage \\ (lost//total)} & \thead{Commands \\ sent N.} &\thead{ Message \\  Delay \\Average [s]} & \thead{ Message \\ Delay \\ Variance [s]}  \\
\hline
Test 1 &10\% &62 & 1.88 & 0.13 \\
Test 2 &14\% &22 & 1.83 & 0.11\\
%Test 3$^{\mathrm{a}}$ & x & x & x & x\\
Test 4 &21\% &58 & 2.04 & 0.17 \\
Test 5 &13\% &37 & 1.85 & 0.09\\
\hline
% \multicolumn{4}{l}{$^{\mathrm{a}}$ Data of test 3 were not recorded}
\multicolumn{4}{l}{$^*$ Data of Test 3 were not recorded}

\end{tabular}
\label{table_statistics}
\end{center}
\end{table}

\begin{figure}[!ht]
\includegraphics[width=8.0cm]{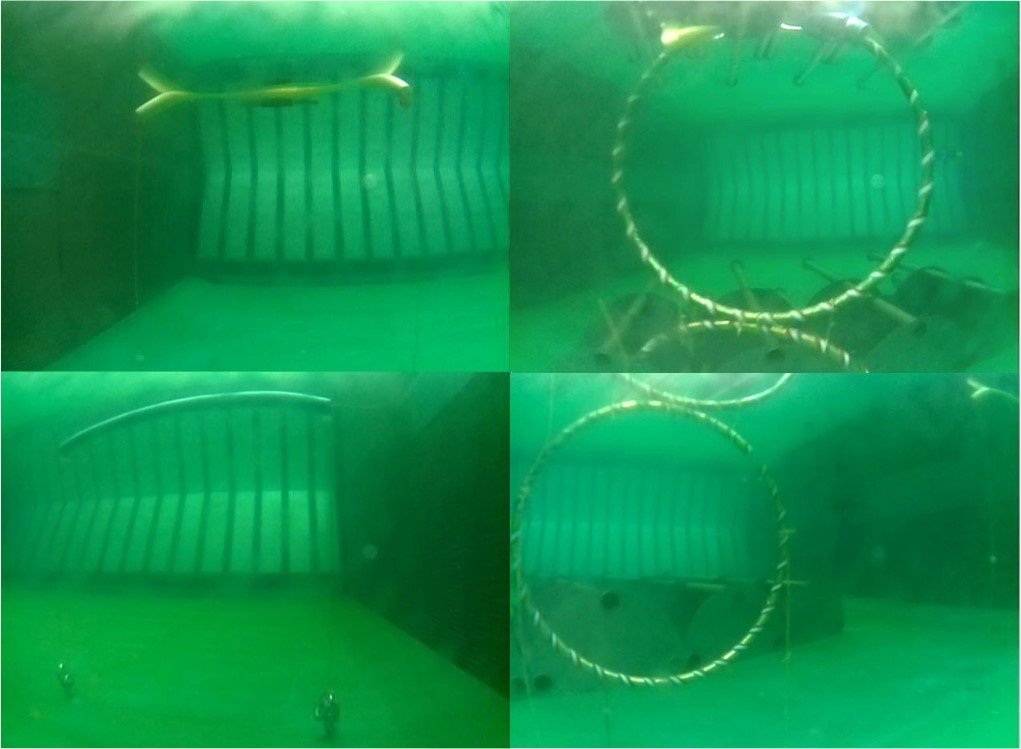}
\centering
\caption{View of each of the gates before being traversed, captured from the frontal camera of the AUV. The gates are ordered left to right, top to bottom.}
\label{fig:onboard}
\end{figure}

In fig.\ref{fig:onboard} is shown the view of the targets from the vehicle frontal camera, collected during the test.

\section{Discussions and Conclusions}
In this work a multi-targets' inspection and monitoring use case has been proposed, in a teleoperation scenario, based on the HSURF multi-platforms  system. The main actors of the system are 1) an operator remotely based, 2) a floating platform (surface vehicle) and 3) a robotic fish (underwater robot). The teleoperation relies on acoustic underwater communication channels  and on wifi as aerial communication. While aerial communication does not produce strong constraints, acoustic communication poses many challenges, due to its well-known intrinsic characteristics and limitations.  
% Discuss results.

The results here presented prove that our method retains an appropriate level of control for simple operations, despite the communication delay, very limited bandwidth  and relative high loss rate, due to the acoustic modem. The current architecture sends commands at fixed intervals, which proved to be effective in the case presented, but could be improved by only sending messages when a command variation is required. Similarly, we have observed that, while the magnetometer was not accurate at estimating the North, and the value could change significantly depending the position relatively to the pool, it allowed for a fairly stable control of the heading, where the operator could add adjustments whenever needed. The expectation is that both magnetic heading and data loss will improve in the open-water, while a code optimization can significantly drop the delay time. 
Nevertheless, the system proved to be reliable for teleoperation via the acoustic channel.

% Consideration about scaling on different robots
The choice of subdividing the control between the operator and onboard is related to the large discrepancy between the fish dynamics and the transmission and feedback time. Currently, the feedback control is based on direct observation from the operator, which adds minimal delay. In an applied scenario, the delay should account for the communication delay back to the operator. This will be included in the future investigation, as the method of returning usable feedback to the operator must be properly investigated. 

Regardless of which type of feedback is implemented, even the current delay for command transmission only was high enough that direct control of the actuators from the operator was not feasible. By delegating depth and heading keeping to the robot, a much higher degree of delay was tolerable. Ideally, the Forward-Backward thrust command should be substituted with movement increments or speed command, but this requires expensive and bulky position or velocity sensors (Long Base Line localization, Doppler Velocity Logger ...) that were not integrated onboard by design. As experienced in preliminary tests, directly controlling the rotation rate (angular velocity) resulted in uncontrolled behavior. On the contrary, with the approach presented, the heading was controllable even in presence of significant magnetic anomalies. 

The HSURF-fish represents, in its design, to be extremely challenge for teleoperation, due to its small size, limited sensors and fast dynamics. Any other class of AUV featuring acoustic communications, will most likely have a slower dynamic and better sensing, which in turn will further enhance the performance of our approach. 

% Seatest preliminary
A preliminary qualitative test in open water has also been successfully performed. The test highlighted the impact of water currents to the operations, but proved that the method allowed a reasonable degree of control of the robotic fish. Such tests were only aimed at confirming the preservation of functionality when in open water. We envision to reconstruct a  setup similar to the one used in the pool to reconfirm and build upon the existing results. The main challenge will be handling the factor of currents and waves' disturbance, which were not present during pool tests. 

% Consideration about multiAUVs 
The work presented can also be framed within a larger scope of the HSURF project, controlling the multi-robots' system. The teleportation aspect in particular is envisioned for a leader-follower approach: the leader will be controlled by the operator, using the method above, while the rest of the AUVs are autonomous and use their relative positions  as input to control their trajectories. In this sense, the teleoperation becomes a building block of a more complex control system.

Furthermore, the ability to directly control the single robot acts as a fail-safe or fine correction tool when operating the full swarm: while from a mission point of view, the loss of one or few AUVs is deemed acceptable, as the mission can negotiate the loss, from a practical point of view is necessary to have retrieval strategy to regain control of those AUVs. 

Regarding the feedback return to the operator, different approaches  are being considered: trilateration through acoustic allows for a general localization, while the most promising approach is the use of video streaming by a tethered sinker: The advantage to a traditional ROV are the possibility of having a third person point of view, especially useful in precise navigation, and the reduction to a single tether in the water. A sinker will not be required precise control to maintain visual contact, while AUVs are not affected by the presence of tethers.  

In a mission scenario, AUVs and Floater-sinker are released in the water. The sinker is deployed at a depth similar to the AUVs and through a set of cameras streams a 360 degree video back to the floater via tether. The video stream is then forwarded to the operator wireless from the floater. The operator is then able to select the desired AUV and send control commands through the acoustic modem mounted on the floater/sinker, and monitor the progress from the point of view of the sinker, while adjusting the point of view by moving the floater and adjusting the sinker depth. The operator can then command a different agent by changing the AUV addressed acoustically. Other than the direct control presented in this paper, we envision a set of different autonomous behaviors (such as station keeping or following another agent) that can be selected and triggered by an operator command, allowing a degree of control on each agent of the multi-robot system.

%  The full paper will contain a more quantitative experimentation of the teleportation scheme proposed. We will use the configuration presented above, and install multiple fixed buoys and suspended targets acting as way-point. 
% %An operator will rely on direct visual contact to guide the robot close to each way-points. 
% Proximity sensors and onboard camera can then be used to estimate the distance reached from the targets.A schematic scene sent via acoustic modem to the operator is in the pipleine as feedback for teleoperation. Mean distance, number of targets reached and overall mission time  will represent a metric of quality for the method proposed.

\section{Acknowledgments}
The authors acknowledge the project “Heterogeneous Swarm of Underwater Autonomous Vehicles” funded by the Technology Innovation Institute (Abu Dhabi)  developed with Khalifa University  (contract no. TII/ARRC/2047/2020).

\bibliographystyle{IEEEtran}
\bibliography{references}

\begin{thebibliography}{10}
\providecommand{\url}[1]{#1}
\csname url@samestyle\endcsname
\providecommand{\newblock}{\relax}
\providecommand{\bibinfo}[2]{#2}
\providecommand{\BIBentrySTDinterwordspacing}{\spaceskip=0pt\relax}
\providecommand{\BIBentryALTinterwordstretchfactor}{4}
\providecommand{\BIBentryALTinterwordspacing}{\spaceskip=\fontdimen2\font plus
\BIBentryALTinterwordstretchfactor\fontdimen3\font minus
  \fontdimen4\font\relax}
\providecommand{\BIBforeignlanguage}[2]{{%
\expandafter\ifx\csname l@#1\endcsname\relax
\typeout{** WARNING: IEEEtran.bst: No hyphenation pattern has been}%
\typeout{** loaded for the language `#1'. Using the pattern for}%
\typeout{** the default language instead.}%
\else
\language=\csname l@#1\endcsname
\fi
#2}}
\providecommand{\BIBdecl}{\relax}
\BIBdecl

\bibitem{oceandecade}
``{OceanDecade} {United Nations} decade of ocean science and sustainable
  development,'' \url{https://oceandecade.org/}, accessed: 2023-01-13.

\bibitem{sudevan2022multisensor}
V.~Sudevan, N.~Mankovskii, S.~Javed, H.~Karki, G.~De~Masi, and J.~Dias,
  ``Multisensor fusion for marine infrastructures’ inspection and safety,''
  in \emph{OCEANS 2022, Hampton Roads}.\hskip 1em plus 0.5em minus 0.4em\relax
  IEEE, 2022, pp. 1--7.

\bibitem{iacoponi2022h}
S.~Iacoponi, G.~J. Van~Vuuren, G.~Santaera, N.~Mankovskii, I.~Zhilin, F.~Renda,
  C.~Stefanini, and G.~De~Masi, ``H-surf: Heterogeneous swarm of underwater
  robotic fish,'' in \emph{OCEANS 2022, Hampton Roads}.\hskip 1em plus 0.5em
  minus 0.4em\relax IEEE, 2022, pp. 1--5.

\bibitem{iacoponi2022iop}
S.~Iacoponi, M.~Hanbaly, A.~Infanti, B.~Andonovski, N.~Mankovskii, I.~Zhilin,
  F.~Renda, C.~Stefanini, and G.~De~Masi, ``Heterogeneous underwater swarm of
  robotic fish: Behaviour and applications,'' in \emph{to appear in
  International Conference on Embodied Intelligence.}\hskip 1em plus 0.5em
  minus 0.4em\relax IOP, 2023.

\bibitem{app10041256}
\BIBentryALTinterwordspacing
J.~González-García, A.~Gómez-Espinosa, E.~Cuan-Urquizo, L.~G.
  García-Valdovinos, T.~Salgado-Jiménez, and J.~A.~E. Cabello, ``Autonomous
  underwater vehicles: Localization, navigation, and communication for
  collaborative missions,'' \emph{Applied Sciences}, vol.~10, no.~4, 2020.
  [Online]. Available: \url{https://www.mdpi.com/2076-3417/10/4/1256}
\BIBentrySTDinterwordspacing

\bibitem{Campagnaro2015SimulationOA}
F.~Campagnaro, F.~Guerra, F.~Favaro, V.~S. Calzado, P.~A. Forero, M.~Zorzi, and
  P.~Casari, ``Simulation of a multimodal wireless remote control system for
  underwater vehicles,'' \emph{Proceedings of the 10th International Conference
  on Underwater Networks \& Systems}, 2015.

\bibitem{7094391}
F.~Campagnaro, F.~Favaro, P.~Casari, and M.~Zorzi, ``On the feasibility of
  fully wireless remote control for underwater vehicles,'' in \emph{2014 48th
  Asilomar Conference on Signals, Systems and Computers}, 2014, pp. 33--38.

\bibitem{jmse8100736}
\BIBentryALTinterwordspacing
F.~Campagnaro, A.~Signori, and M.~Zorzi, ``Wireless remote control for
  underwater vehicles,'' \emph{Journal of Marine Science and Engineering},
  vol.~8, no.~10, 2020. [Online]. Available:
  \url{https://www.mdpi.com/2077-1312/8/10/736}
\BIBentrySTDinterwordspacing

\bibitem{muscolo_marcheschi_fontana_bergamasco_2021}
G.~G. Muscolo, S.~Marcheschi, M.~Fontana, and M.~Bergamasco, ``Dynamics
  modeling of human–machine control interface for underwater teleoperation,''
  \emph{Robotica}, vol.~39, no.~4, p. 618–632, 2021.

\bibitem{Cardenas2016}
E.~F. Cárdenas and M.~S. Dutra, ``An augmented reality application to assist
  teleoperation of underwater manipulators,'' \emph{IEEE Latin America
  Transactions}, vol.~14, no.~2, pp. 863--869, 2016.

\bibitem{padrao2023towards}
P.~Padrao, J.~Fuentes, T.~Kaarlela, A.~Bayuelo, and L.~Bobadilla, ``Towards
  optimal human-robot interface design applied to underwater robotics
  teleoperation,'' \emph{arXiv preprint arXiv:2304.02002}, 2023.

\bibitem{Akyildiz2004}
I.~F. Akyildiz, D.~Pompili, and T.~Melodia, ``Challenges for efficient
  communication in underwater acoustic sensor networks,'' \emph{SIGBED Rev.},
  vol.~1, no.~2, p. 3–8, Jul. 2004.

\bibitem{fossen2011handbook}
T.~I. Fossen, \emph{Handbook of marine craft hydrodynamics and motion
  control}.\hskip 1em plus 0.5em minus 0.4em\relax John Wiley \& Sons, 2011.

\end{thebibliography}
\end{document}